\documentclass[preprint,12pt]{article}

\usepackage{graphicx}
\usepackage{color}
\usepackage{amssymb}
\usepackage{caption}
\usepackage{algorithm}
\usepackage{algorithmic}
\usepackage{verbatim}
\usepackage{eurosym}

\usepackage{amsmath}

\title{Robust human position estimation in cooperative robotic cells
	
	\thanks{\textit{\underline{Citation}}: 
		\textbf{Amorim, A., Guimares, D., Mendonca, T., Neto, P., Costa, P., Moreira, A. P. (2021). Robust human position estimation in cooperative robotic cells. Robotics and Computer-Integrated Manufacturing, 67, 102035. https://doi.org/10.1016/j.rcim.2020.102035 }} 
	
}

\author{
	A. Amorim, D. Guimarães, T. Mendonça, P. Neto, P. Costa, A.P. Moreira \\
	INESC TEC, University of Porto, Portugal \\
	University of Coimbra, Portugal \\
}

\begin{document}
	\maketitle

\begin{abstract}
Robots are increasingly present in our lives, sharing the workspace and tasks with human co-workers. However, existing interfaces for human-robot interaction / cooperation (HRI/C) have limited levels of intuitiveness to use and safety is a major concern when humans and robots share the same workspace. Many times, this is due to the lack of a reliable estimation of the human pose in space which is the primary input to calculate the human-robot minimum distance (required for safety and collision avoidance) and HRI/C featuring machine learning algorithms classifying human behaviours / gestures. Each sensor type has its own characteristics resulting in problems such as occlusions (vision) and drift (inertial) when used in an isolated fashion. In this paper, it is proposed a combined system that merges the human tracking provided by a 3D vision sensor with the pose estimation provided by a set of inertial measurement units (IMUs) placed in human body limbs. The IMUs compensate the gaps in occluded areas to have tracking continuity. To mitigate the lingering effects of the IMU offset we propose a continuous online calculation of the offset value. Experimental tests were designed to simulate human motion in a human-robot collaborative environment where the robot moves away to avoid unexpected collisions with de human. Results indicate that our approach is able to capture the human\textsc{\char13}s position, for example the forearm, with a precision in the millimetre range and robustness to occlusions.
\end{abstract}


%%\linenumbers

\section{Introduction}
\label{S:1}
Robots are becoming our workmates. They have an increasingly central role in the companies\textsc{\char13}s competitiveness, with subsequent positive impact in jobs creation and social welfare. The paradigm for robot usage has changed in the last few years, from an idea in which robots work autonomously to a scenario in where robots collaborate with human beings. This means taking advantage of the best abilities of each partner, by exploring the coordination and cognitive capabilities of humans and the accuracy and capacity to produce monotonous tasks of robots. Thus, robots and humans have to understand and interact in a natural way with each other.\par

The problem is that existing interfaces for human-robot interaction / collaboration (HRI/C) have limited levels of intuitiveness to use and safety is still a major concern when humans and robots share the same dynamic and unstructured workspace. The number of studies reporting results using real robots (not in simulation) operating in real environments is still limited. Many times, such challenges arise due to the lack of a reliable estimation of the human pose in space (human tracking). The human pose in each instant of time is the primary input to calculate the human-robot minimum distance (required for safety and collision avoidance) and HRI/C featuring machine learning algorithms classifying for example human behaviours / gestures. Existing collision avoidance solutions are based on workspace monitoring, when a human is detected the robot stops or reduces velocity. Safety requirements for collaborative robots are defined by the standard ISO 10218 and the technical specification TS 15066 \cite{safetysaenz}. Reliable human tracking allows to define proper human-robot safety distances and reactive collision avoidance methods \cite{SAFEEA201933,SAFEEA2019278}. Moreover, human tracking is also required for the pattern recognition of human behaviours and gestures, which demonstrated to be effective and intuitive human-robot interfaces \cite{TSAROUCHI20161}.\par

Several approaches for physical and cognitive HRI/C have been investigated, using different interaction technologies, manipulators, learning methods and safety strategies \cite{doi:10.1177/0278364909343970, SCHMIDT2014711}. A major goal of current research is to study novel methodologies that help ordinary users to intuitively interact and collaborate with a robot in safety. Multiple sensors (vision, laser scanners, inertial, magnetic, etc.) have been used to achieve reliable human tracking and by this way guaranteeing a more effective, safe and intuitive HRI/C.\par

In this study we aim to improve existing methods for human position tracking relying on 3D vision sensing and low-cost IMUs. These methods already provide a relatively good accuracy in human position tracking. However, error arises from sensor occlusions, which in the context of HRI/C, can be a safety hazard. By introducing an IMU to the system, we are able to compensate the motion capture with an estimation derived from its data, albeit less precise. We also address the inherent problems of low-cost IMUs such as their low sensitivity and time varying offsets affecting their measurements. Using higher-cost IMUs these issues are minimized leading to a trade-off between component cost and the quality of its measurements. The proposed methodologies are validated in different experiments and a use case.\par

This paper\textsc{\char13}s main objective is to develop a system capable of performing reliable human position tracking, which may be later applied to perform collaborative tasks involving robots and humans. To address this problem, the following objectives are considered:

\begin{itemize}
	
	\item Track human body\textsc{\char13}s 3D position with high precision while having the possibility of  tracking errors due to occlusions;
	\item Relate the different coordinate frames (global, camera, IMU) in such a way that it allows flexibility for the IMU placement on the operator body;
	\item Acquire IMU data that portrays the human operator movement in order to maintain a reliable tracking when occlusion errors occur.
	
\end{itemize}

\subsection{Related work}

HRI/C is critical for the acceptance of collaborative robots that work side-by-side with human co-workers \cite{Simao2017UnsupervisedStream}, especially when they are operating in small lot size production. The development of intuitive HRI/C interfaces has been enormous in the last decades \cite{Burke2015PantomimicInteraction}, so that today robot assistants are present in the most diverse manufacturing companies and domains. However, they tend to heavily rely on controlled environments, demonstrate poor autonomy and limited collaborative capabilities \cite{Pedersen2016RobotDeployment}.

Different sensors have been combined to achieve accurate and reliable human tracking, from inertial to magnetic to vision-based sensors. Magnetic tracking sensors measure precise body motion, but they are very sensitive to magnetic noise and need to be attached to the human body. Inertial sensors are wearables that are relatively cheaper, allow tracking independently of the body orientation and can be used in cluttered environments (no occlusions). On the other hand, they lack in accuracy in long term usage, especially due to drift effect. A major advantage of using vision-based systems is the non-intrusive character of this technology. However, they have difficulty producing robust information when facing cluttered environments and occlusions. Some vision-based systems are view dependent, require a uniform background and illumination. Looking to literature, the interaction technologies for HRI/C are mostly based on vision (including laser scanners) and wearable sensors (magnetic, inertial, electromyographic, etc.) \cite{Ordonez2016DeepRecognition, Wolf2013Gesture-basedBioSleeve}. Recent advances demonstrated that in unstructured environments reliable data related to human position can be obtained with hybrid solutions. Deep learning allowed for example to classify gestures from incomplete 3D vision data \cite{Wu2016DeepRecognition}. However, the recognition of human behaviours and gestures is still a challenge, especially when using vision-based systems which are natural (no need to attach sensors to the body) but with limited reliability due to occlusions, varying light conditions and changing backgrounds \cite{Pinto2014UnsupervisedSystem}. Researchers have been studying background invariant hand gesture detection using OpenPose library integrated with Microsoft Kinect V2 data to obtain a 3D estimation of the human skeleton \cite{MAZHAR201934}. While background is not a problem for wearable sensors, the repeatability of IMUs has been studied, demonstrating demonstrating increasing accuracy and repeatability \cite{wwws}. A human motion tracking approach combining a mobile robot (equipped with a laser scanner) and an inertial motion capture system is proposed in \cite{6094430}. The mobile robot is used to anchor the pose estimates of the human which is wearing a motion capture suite quipped with 17 Xsens IMUs to estimate the body posture. The system captures the motion of a human in large areas (outdoor) fusing data from laser and IMUs for more accurate tracking. The study presents the trajectory of the human in meters scale (outdoor) but the entire skeleton pose is not detailed. Depth data from a Kinect camera have been used to calculate distances between humans and reference points on robots \cite{6225245}. Depth cameras have also been studied to improve the monitoring in a human-robot collaborative environment \cite{SCHMIDT2014711}. This is important since the lack of sensors reporting reliable data is a major problem in this kind of applications. The re-construction of the environment surrounding the robot based on two laser scanners placed on the robot mobile platform and a Kinect sensor located on its torso is proposed in \cite{KOUSI20191429}. Optical sensors have been used for workspace monitoring featuring human collision risk detection \cite{NIKOLAKIS2019233}. Researchers also proposed the monitoring of human-robot coexistence using depth sensors and laser scanners to estimate the relative human-robot distance \cite{MAGRINI2020101846}. Such a distance is the input for a layered control architecture to modify in real time the robot behaviour. Nevertheless, the reported accuracy for human pose estimation is the one provided by the depth sensors, which is not robust to occlusions. In a different approach, it is proposed real-time monitoring aiming to secure the minimum protective distance between humans and robots recurring to depth-sensing based models and interactive augmented reality \cite{HIETANEN2020101891}. Results report a good performance regarding safety and ergonomics. However, the problem of occlusions is not addressed. Wearable inertial sensors have been used for human motion tracking \cite{netokalman}, not requiring external cameras or markers. They can be used in both outdoor and indoor environment, with no light restrictions nor suffering from occlusions. Nevertheless, drift is a major problem associated to inertial-based sensors, especially over long periods of time. Some authors propose to correct the estimated quantities (for example the position of the body with respect to a coordinate system not fixed to the body) by updating these quantities based on biomechanical characteristics of the human body, detection of contact points of the body with an external world and adding other sensors to the system. IMUs have been used to measure the orientation of the human hand combined with a Camshift algorithm to track the human hand from a 3D camera \cite{DU201693}. Kalman filter and particle filter are used to estimate the orientation and the location of the human hand. Results indicate an error of about 1 mm in 2D and 2.5 mm in 3D. These errors were obtained in relatively controlled environment and in a small volume. The robustness to occlusions is not considered. 

A recent study revises the existing approaches on collaborative robots with focus on safety requirements and safety assurance, highlighting the importance of safety and human-robot interfaces for future research in collaborative robotics \cite{BI2021102022}. The challenging aspects of close collaboration between humans and robots, and the required ability for the robot to adapt its behaviour to the human-in-the-loop have been studied \cite{Flacco2015ControlSpace}. Safe HRI/C requires monitoring the workspace to guarantee a minimum separation distance where the reliable tracking of human position is critical. An interesting study defines four safety strategies for workspace monitoring and collision detection where human tracking is required \cite{doi:10.1080/0951192X.2016.1268269}. Some authors report that to guarantee safety the robot motion must be restricted to a safe region \cite{Kimmel20126DReduction}. This is a major constraint to a collaborative scenario which can be attenuated using reactive control strategies \cite{Haddadin2008CollisionInteraction}. Existing effective safety solutions are based on workspace monitoring strategies where when a human is detected the robot stops or reduces the velocity \cite{SCHMIDT2014711}. Motivated by the fact that current distance-based collision-free human-robot collaboration only ensures human safety but not assembly efficiency, researchers proposed a collision-free human-robot collaboration system based on context awareness \cite{LIU2021101997}. The proposed system plans robot paths while reaching target positions in time. Human operators’ assembly poses are recognised using transfer learning requiring low computational expenses. Different sensors were tested to recognise human poses, namely sensors for hand poses capturing (EMG-based MYO and Leap Motion) and sensors to detect the human poses of the whole (Kinect depth sensor and 2D cameras).  

Several studies in literature address human tracking combining different sensor data and methods. Owing to the nature of the problem, the number of studies dealing with occlusions and motion continuity is still limited or based on highly expensive sensors. Our proposed solution aims to fill the gap that exists in human tracking continuity when occlusions occur. This is achieved by combining 3D vision and IMUs data, implementing a novel IMU compensating tracking method that demonstrated to be efficient for the human-robot interactive process in a shared environment.

\section{Methodology}

As reported in previous section the reliable estimate of human positions in space is still a challenging process due to several factors. In this paper we propose to address these challenges and limitations of existing systems, demonstrating that it is possible to track human motion while performing a human-robot collaborative task. To accomplish that, our method passes by estimating the pose and velocity of human body limbs by fusing data from a 3D vision sensor and a set of IMUs placed in human body limbs.

The human tracking is achieved using a 3D vision sensor (Optitrack) but other 3D sensors can be used. This system can track a human defined body in a maximum of 12 individuals trough strategically placed infrared reflective spheres and a set of infrared cameras. However, in certain positions, some of the spheres might get occluded which leads to a temporary gap in the tracking system where no information about the body\textsc{\char13}s position is given. In these instants the information given by the IMU attached to the human body is processed in order to maintain the body's tracking.\par

By combining the position measured by the Optitrack system with the acceleration data given by the IMU, which we can then use for position estimation, we obtain a continuous and reliable (robust to occlusions) position tracking system. Such information can be used in different HRI/C applications, for example to estimate the human-robot minimum distance for safety purposes and/or to understand human behaviours/gestures pursuing natural interfacing with robots.\par

In order to mitigate the lingering effects of the IMU offset, we propose a continuous, online calculation of the offset value. When the main tracking system is running exempt of errors, we can compute the IMU offset with high precision, whereas, on the event of a fault which triggers the pose estimation from the IMU data, we adjust the last calculated offset value accordingly. This is a major issue and the development of reliable methodologies to tackle it is key when using low-cost IMUs. Both the 3D vision sensors and the IMUs are not safety certified devices to be directly used in industry without proper risk analysis for each specific human-robot collaborative application.

\subsection{System architecture}
\begin{figure}
    \centering
    \includegraphics[width=0.8\linewidth]{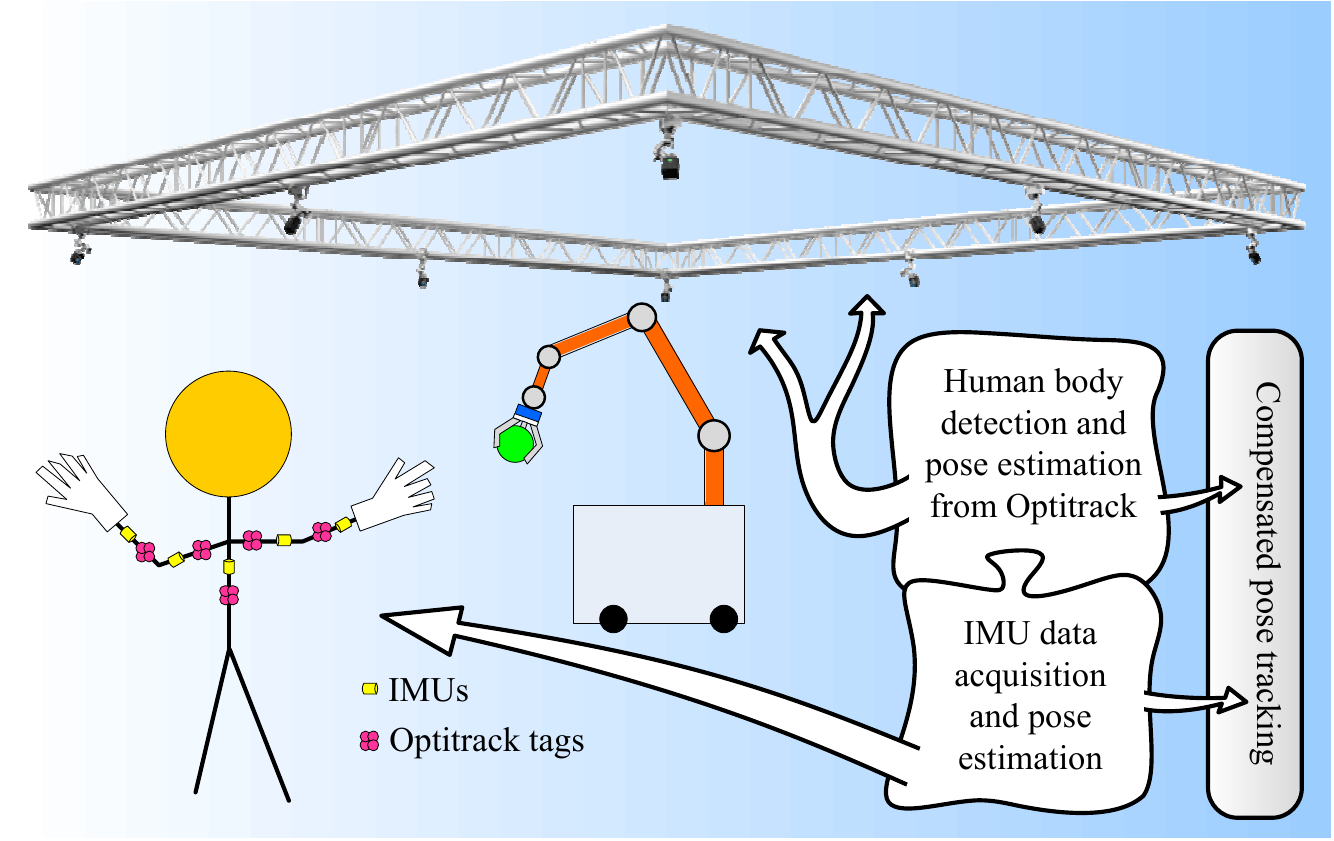}
    \caption{Architecture of the proposed human tracking system.}
    \label{fig:archi}
\end{figure}

\begin{figure}
	\centering
	\includegraphics[width=1\linewidth]{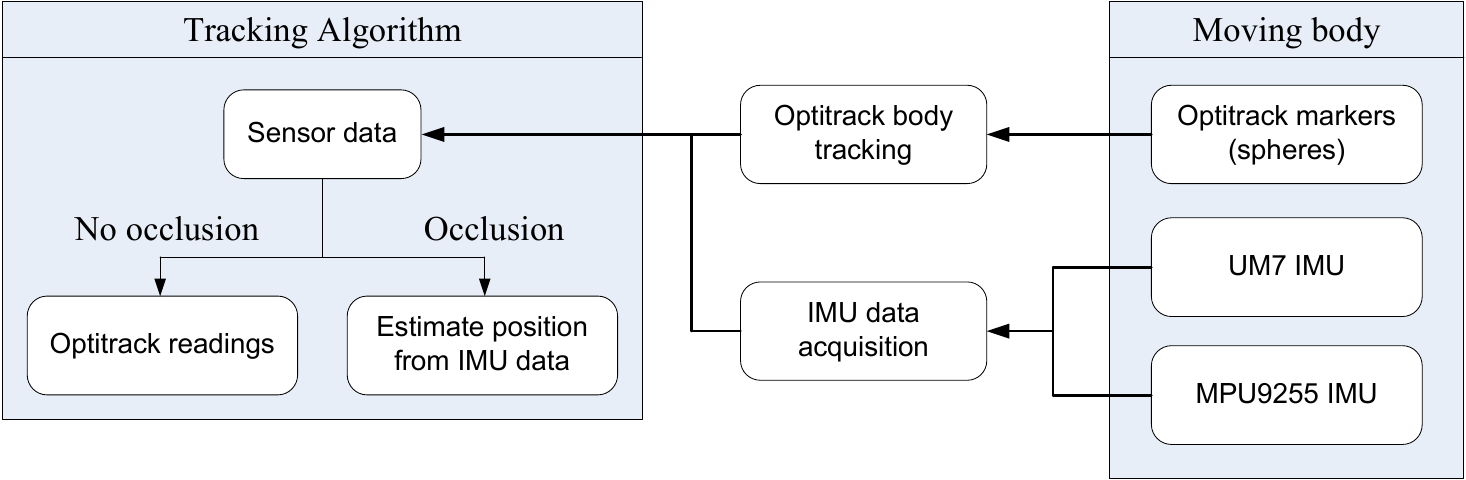}
	\caption{Components interaction featuring both hardware and software.}
	\label{fig:archi22}
\end{figure}

Fig. \ref{fig:archi} shows the overall architecture of the system which is decomposed in three modules. The first module is dedicated to IMU data acquisition, processing and extraction of position and orientation estimations. The second module provides the main human body detection and pose estimation using the Optitrack system. The last module ensures the tracking continuity of the moving body, compensating any failures (occlusions) that might occur during the main detection provided by the second module (3D vision) using the data being acquired by the first module (IMUs).The diagram in Fig. \ref{fig:archi22} shows the connections between the different elements featuring both hardware and software components.

\subsection{IMU data}
The IMU is composed by an accelerometer, gyroscope and magnetometer. The accelerometer data, trough integration, outputs position estimation. However, the accelerometer offset, which stems from the IMU\textsc{\char13}s manufacturing process and may be variable throughout its usage, heavily hinders the position estimation trough double integration. The offset\textsc{\char13}s impact is so significant that low-cost IMU are hardly ever used in this type of applications. In this paper, we tackle this problem. The gyroscope and magnetometer, either alone or combined, provide orientation data.

The IMU\textsc{\char13}s data are measured according to its own coordinate frame, which will almost certainly be different than the main tracking source\textsc{\char13}s frame, which in turn might be different from the global frame. The transformations between these coordinate frames are needed in order to correctly interpret the data that we measure. In turn, this may present a constraint around the IMU mounting on a human operator. With that in mind, we were able to easily obtain the transformations, without previously knowing any physical information, using the Kabsch algorithm \cite{kabsch}. By collecting sets of data points from both sources (Optitrack and IMU) in different approximately orthogonal positions along all 3 axes of the global frame, the aforementioned algorithm computes the optimal rotation matrix between the 2 sets of paired points. In our application, this only needed to be done once since the IMU frame was fixed in relation to the moving body frame.

%possivelment meter tabela com resultados do Kabsch

\subsection{3D vision sensor}
As previously mentioned, the Optitrack system was used as the main source of the body\textsc{\char13}s pose data. The system setup used six FLEX3 cameras that were able to track a moving body, properly covered in reflective spheres, on a bounded workspace with very high precision. However, this system requires that at least two cameras detect the spheres at every instant, which means that there is a possibility to occlude one or more cameras, leading to a tracking error. Since our objective is to track human body limbs, this possibility gains magnitude. In these instants, where the Optitrack system is not able to measure the pose of the body, we rely on the IMU readings to estimate positions.

\subsection{IMU compensated tracking}

The algorithm we developed for IMU compensated tracking assumes that as the pose tracking is taking place, a buffer is storing the last $N$ position readings. When no occlusion is detected, a 3rd order polynomial (Equation~\ref{eq:poly}) is adjusted to such readings, allowing the extraction of the expected acceleration for an instant $t$ (Equation~\ref{eq:acc}).

\setlength{\abovedisplayskip}{0pt} \setlength{\abovedisplayshortskip}{0pt}
\begin{equation}\label{eq:poly}
    ymeas(t) - y_{0} = v_{0} \cdot t + a_{0}\cdot \frac{t^{2}}{2} + a_{a}\cdot \frac{t^{3}}{6}
\end{equation}

\begin{equation}\label{eq:acc}
    a(t) = a_{0} + a_{a}\cdot (t-t_{0})
\end{equation}

\noindent Where $ymeas(t)$ is the measured position and $a(t)$ is the expected acceleration at instant $t$, $y_{0}$, $v_{0}$, $a_{0}$ and $a_{a}$ are the position, velocity, acceleration and jerk on the $t-N$ position reading, respectively. These equations assume that the acceleration varies linearly with time, as opposed to the more traditional methods which assume constant acceleration. For simplicity\textsc{\char13}s sake, this set of equations illustrates the procedure for a single coordinate, along the y axis. The expansion to the remaining coordinates follows a similar process. Moreover, the equations represent a continuous motion. In order to estimate the values of $v_{0}$, $a_{0}$ and $a_{a}$, we formulate a Least Squares Estimation problem with two discrete set of points (Equation~\ref{eq:y1} and Equation~\ref{eq:vel2}).

\begin{equation}\label{eq:y1}
    Y =
    \begingroup % keep the change local
    \setlength\arraycolsep{4pt}
    \begin{bmatrix} 
    y[k+1] - y[k] \\ y[k+2] - y[k] \\ \dots \\ y[k+N] - y[k] 
    \end{bmatrix}
    \endgroup
\end{equation}
\begin{gather}\label{eq:vel2}
    \begingroup % keep the change local
    \setlength\arraycolsep{4pt}
    X = \begin{bmatrix} Ta & \frac{Ta^{2}}{2}  & \frac{Ta^{3}}{6} \\ 2\cdot Ta & \frac{(2 \cdot Ta)^{2}}{2}  & \frac{(2 \cdot Ta)^{3}}{6} \\ \dots &\dots& \dots \\ N\cdot Ta & \frac{(N \cdot Ta)^{2}}{2}  & \frac{(N \cdot Ta)^{3}}{6} \end{bmatrix}\endgroup
\end{gather}

\noindent Where $k=t-N$, and $Ta$ is the sampling time of the Optitrack position readings. We adopted the same procedure to estimate the current velocity $v$ of the moving body (Equation~\ref{eq:vel3}), however, we use the last $Nv$ points where $Nv \approx \frac{N}{4}$. The values of $N$ and $Nv$, as well as the relationship between them, were decided upon trough experimentation. By using a smaller amount of points to estimate the current velocity, in the event of two or more, occlusions happening in such quick succession that there are not enough points to estimate the expected acceleration, at least there is a very high likelihood that we can still estimate the current velocity. 

\begin{equation}\label{eq:vel3}
    v = (X_{1}'\cdot X_{1})^{-1}\cdot X_{1}'\cdot Y
\end{equation}

\noindent Where $X_{1}$ corresponds to the first column of $X$ where $N=Nv$ and $Y$ is the same matrix as in Equation~\ref{eq:y1} with $N=Nv$. The expected acceleration is rotated to the IMU frame and compared to the measured acceleration from the IMU. Their difference corresponds to the offset calculated on the IMU axis:

\begin{equation}\label{eq:off1}
    offset = a_{read} - a(t)
\end{equation}

\noindent Where $a_{read}$ is the value of the acceleration read from the IMU. If we are already in the estimation process and therefore do not have a value for the expected acceleration, we adjust the offset:

\begin{equation}\label{eq:off2}
    offset = offset_{prev} + 0.01 \cdot (a_{read} - a(t) - offset_{prev}) 
\end{equation}

\noindent Where $offset_{prev}$ is the offset in the previous instant. The measured acceleration from the IMU is rotated to the global frame and adjusted according to a gain $K$ (Equation~\ref{eq:rotation}). This value was calculated offline, only once, from the difference between the velocity obtained by differentiating the position readings and integrating the acceleration readings. However, trough experimentation, we verified that, if the absolute value of the expected acceleration was too high, it was best to skip the offset calculation for the instant we are analysing as the calculated value was erroneous. When the previous condition is met, we simply assume that the offset value for the instant is the same as for the previous one.\par

\begin{equation}\label{eq:rotation}
    a_{IMU} = K \cdot R_{body}^{global} \cdot R_{IMU}^{body} \cdot (a_{read} - offset)
\end{equation}

\noindent Where the generic $R_{x}^{y}$ is the rotation matrix from the $x$ frame to the $y$ frame and $a_{IMU}$ is the measured acceleration given by the IMU in the global frame.\par
The position of the moving body is then estimated, assuming a constant acceleration model between two successive readings, by using the measured and calculated values:

\begin{equation}\label{eq:position}
    Yest[t] = Yest[t_{-1}] + v{0} \cdot (t-t_{-1}) + a_{IMU}\cdot\frac{(t-t_{-1})^{2}}{2}
\end{equation}

\noindent Where $Yest(t)$ is the estimated position at instant $t$. 

\begin{algorithm}
\caption{$Tracking Algorithm$}\label{alg:preenchey}
\begin{algorithmic}

\IF{occlusions on last Ncalc measures}
    \STATE {Offset = Previous Offset}
\ELSE
    \STATE{Calculate expected acceleration trough 3rd order polynomial}
    \STATE{Calculate body rotation}
    \STATE{Rotate expected acceleration to IMU frame}
    \STATE{Measure Offset}
        \IF{First}
            \STATE{First = FALSE}
            \STATE{Calculate Offset}
        \ELSE
            \IF{Absolute expected acceleration too high}
                \STATE{Offset = Previous Offset}
            \ELSE
                \STATE{Adjust Offset}
            \ENDIF
        \ENDIF
\ENDIF

\IF{No occlusions on last Nv measures}
    \STATE{Calculate last velocity from last Nv measures}
\ENDIF

\IF{Occlusion right now}
    \STATE{Set initial velocity as last velocity}
    \STATE{Rotate IMU readings to global frame}
    \STATE{Subtract Offset to measured aceleration}
    \STATE{Calculate position trough constant acceleration model and last estimated position}
    \STATE{Estimate orientation}
    \STATE{Update body to global frame rotation based on orientation readings}
    \STATE{Update last velocity}
\ENDIF

\end{algorithmic}
\end{algorithm}

\section{Experiments}

This section describes the setup, methodology and results obtained in the experiments. 
The hardware set was composed by:
\begin{itemize}
    \item 1 MPU9255 IMU (from Waveshare);
    \item 1 UM7 IMU (from CHRobotics);
    \item a set of 6 OptiTrack Flex 3 cameras;
    \item a plastic tube serving as a base structure.
\end{itemize}

\begin{figure}
  \centering
  \includegraphics[width=1\linewidth]{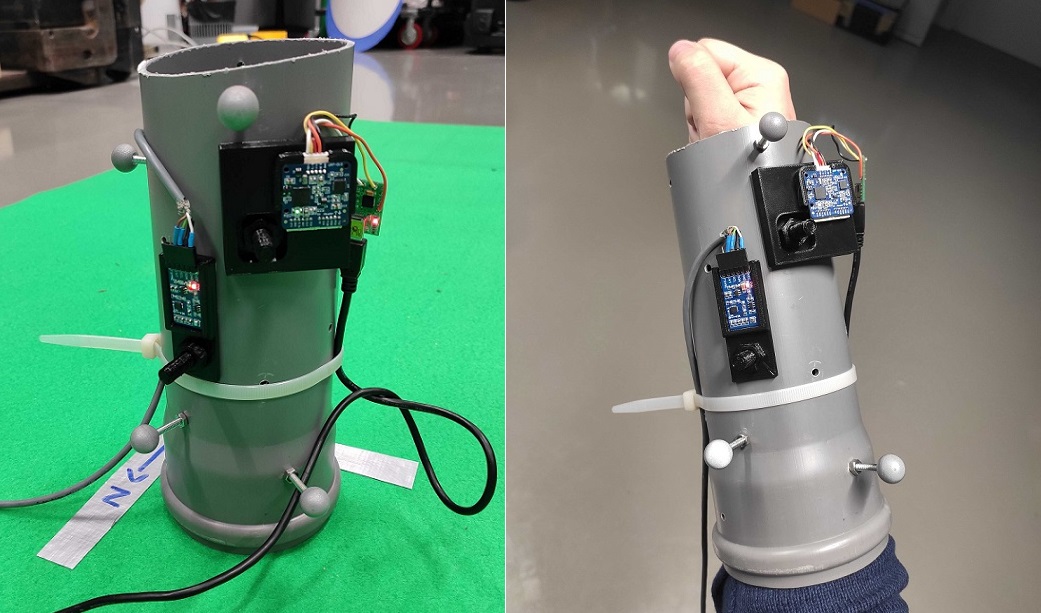}
  \caption{Plastic tube with IMUs and Optitrack reflective spheres (left) and the complete system on the human forearm (right).}
  \label{fig:tubo}
\end{figure}

In our experiments the main objective is to track the forearm of a human. Both the IMU and the Optitrack reflective spheres are attached to the human forearm using a plastic tube, Fig. \ref{fig:tubo}. For bench test purposes, two IMUs were assembled on the tube but we will mainly use and analyse data from the MPU9255 as it is the lowest cost one.

%\begin{figure}[htbp]
%  \centering
%  \includegraphics[width=1\linewidth]{figures/gr2.jpg}
%  \captionof{figure}{Plastic tube on human arm}
%  \label{fig:tubonobraco}
%\end{figure}

\begin{figure}
	\centering
	\includegraphics[width=0.8\linewidth]{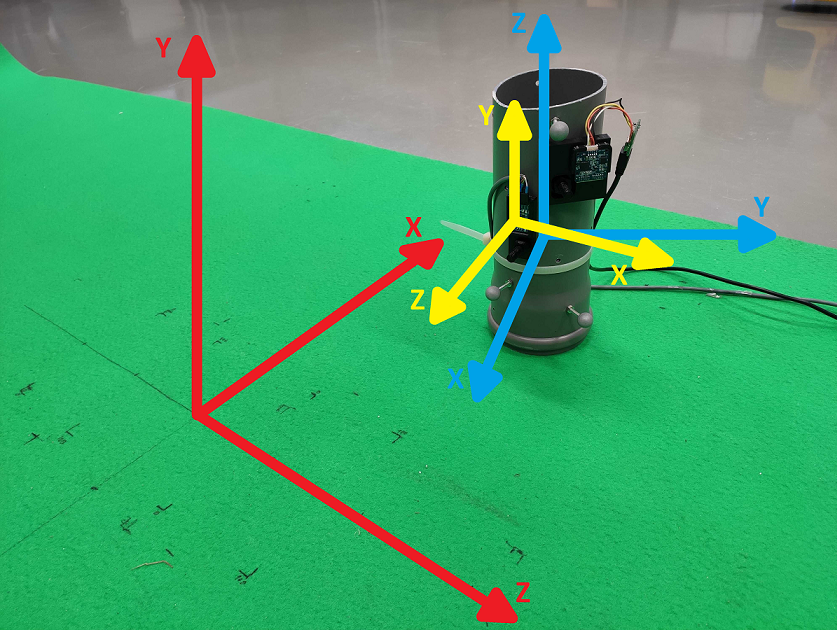}
	\caption{Global frame (red), body frame (blue) and IMU frame (yellow).}
	\label{fig:eixos}
\end{figure}

\begin{figure}
	\centering
	\includegraphics[width=1\linewidth]{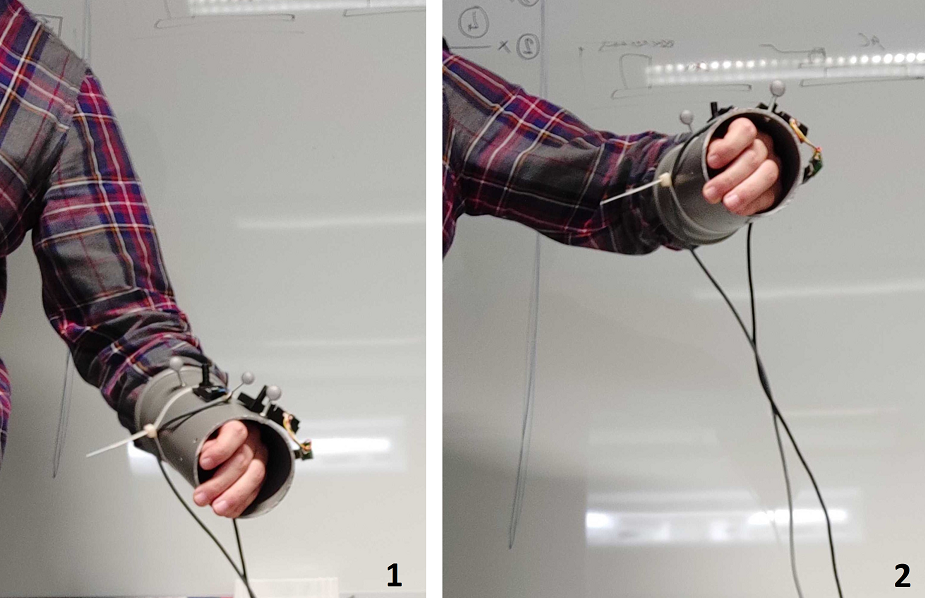}
	\caption{Human movement along the $y$ axis on the global frame, "arm lift".}
	\label{fig:camshot1}
\end{figure}

\begin{figure}
	\centering
	\includegraphics[width=0.9\linewidth]{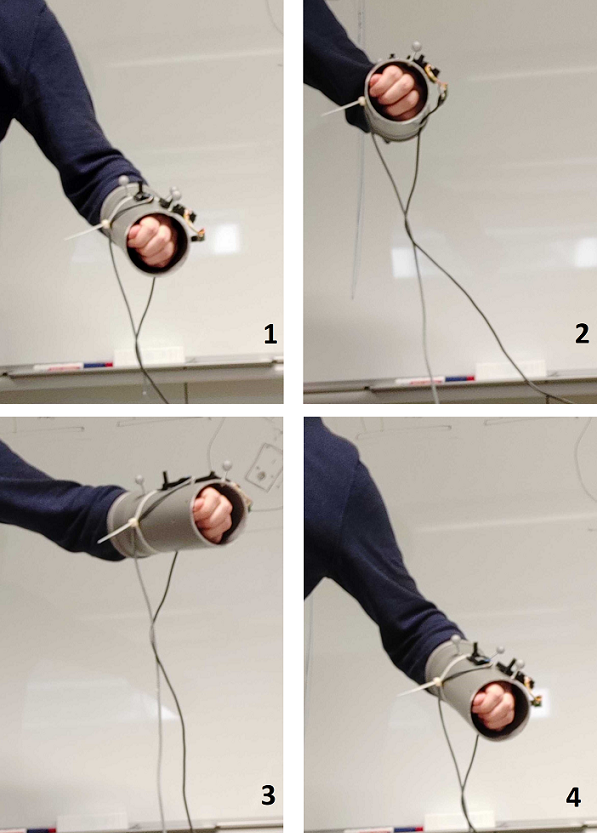}
	\caption{Human movement along multiple axis, roughly circular trajectory.}
	\label{fig:camshot2}
\end{figure}

\begin{figure}
	\centering
	\includegraphics[width=1\linewidth]{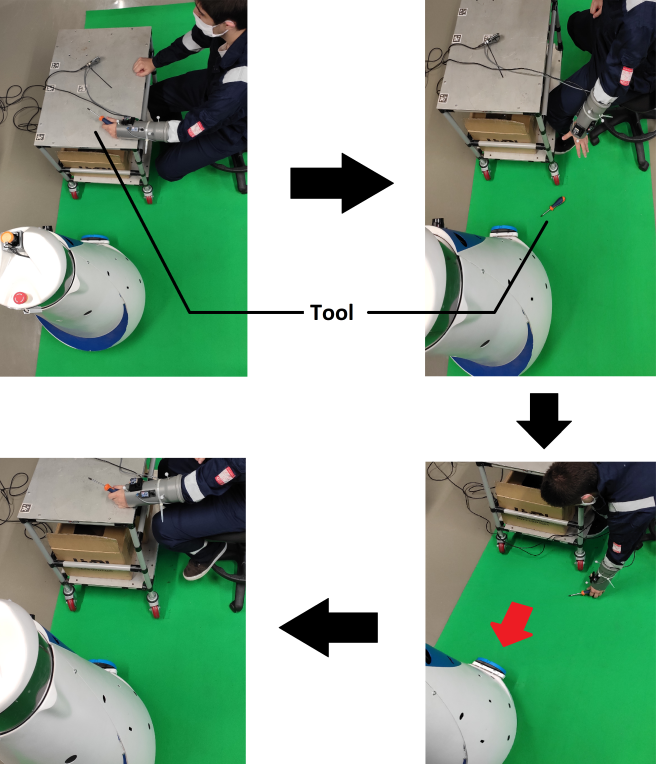}
	\caption{Collaborative use case scenario.}
	\label{fig:fig-case}
\end{figure}

The Optitrack system was assembled with a 6-camera configuration. Owing to the reflective properties of the laboratory\textsc{\char13}s floor, a rug was used to cover it while also visibly demonstrating the system\textsc{\char13}s workspace, which is approximately 1.2x3x2 meters. As per the Optitrack standard global frame, the ground plane is defined by the $x$ and $z$ axis, while the $y$ axis is vertical pointing upwards. The relationship between this global frame, the moving body and the IMU frame is shown in Fig. \ref{fig:eixos}. The plastic tube was used to fix the sensors and demonstrated effective doing that during the experiments. However, this is not a comfortable solution for the humans interacting with a robot. Our ongoing work is investigating the integration of these sensors into smart textile/clothes.   

In a first test we focused on tracking the human forearm movement along the $y$ axis on the global frame, a simple "arm lift", corresponding to an upward and downward movement of translation without rotating the arm, Fig. \ref{fig:camshot1}. Since there is no rotation of the forearm, we could determine the initial rotation from the arm to the global frame and use it throughout the entire test. Like so, we could focus only on the accelerometer data, which is prone to the most significative errors, to validate the position estimation.

Subsequently, in a second test we focus on tracking the human movement along multiple axis, namely a roughly circular trajectory on the $x$ and $y$ global frames, Fig.~\ref{fig:camshot2}. For these forearm motions, on the event of an occlusion, the position in all axes was estimated, as well as the orientation of the body in the global frame using the gyroscope data.

The proposed solution was tested on a use case featuring a shared workspace where the estimated human arm position is the main input that allows the robot to react to avoid collision when the human approaches, Fig. \ref{fig:fig-case}. The human operator, equipped with the necessary sensors on his/her forearm, is performing assembly tasks while sitting next to a stationary robot. The robot is a mobile robot inspecting the human working table. The operator suddenly drops a tool onto the floor, relatively close to the robot. While picking up the tool from the ground, the robot, which is receiving information about the operator’s forearm pose, reacts and moves away to avoid any possible collision. The HRI/C process was enhanced since the operator had more room to pick up the tool without colliding with the robot and by this way feeling safer by increasing its confidence in the shared workspace. Moreover, this use case is based on an important premise that defines HRI/C, the robot adapts itself to the human and not its opposite. During this process, the 3D sensor tracking is briefly occluded by the robot and the operator’s workstation. After picking up the tool, the operator resumes work and the robot returns to its previous position.

\begin{figure}
	\centering
	\includegraphics[width=0.8\linewidth]{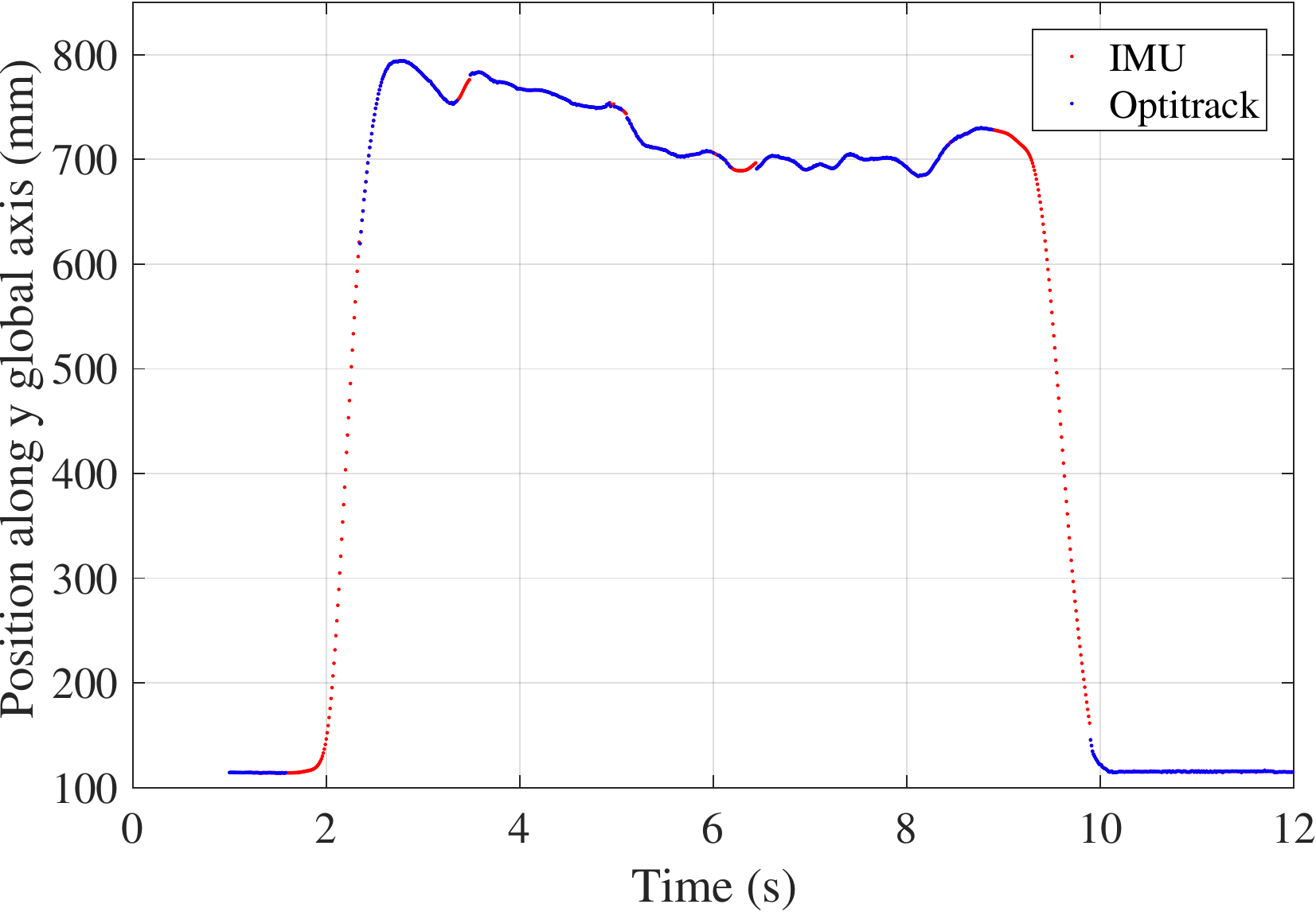}
	\caption{Measured position (blue) and  accelerometer data position estimation (red)}
	\label{fig:grafico}
\end{figure}

\subsection{Results and discussion}
Concerning the test addressing the forearm upward and downward movement along the $y$ axis on the global frame, the gaps along the trajectory, caused by occlusions, are filled by position estimation from the IMU data (MPU9255), Fig.~\ref{fig:grafico} in red. The gaps occur when there is a change in motion, when the forearm goes up and down. We can observe that the estimation based on the IMU data is adequate and fits the expected trajectory, with an approximate error of 15 millimetres in the worst case. Following this testing procedure, similar experiments were conducted to better analyse the lost cost IMU (MPU9255) performance. Two body (human arm) motions were analysed when the occlusion is occurring. In Type 1 the body motion is an upward and downward movement where the occlusion occurs when the arm is going upward or downward. In Type 2 the occlusion occurs when the arm changes direction from upward to downward. Table~~\ref{table:results} shows the collected results. In both experiments the arm is moving at varying speeds and the global axis where the most prominent motion occurred varied as well. The errors we measured varied slightly due to the aforementioned conditions. We can observe a better estimation during Type 1 motions in general, which is especially true considering the last two experiments. This is most likely due to the speed at which the body was moving, since these two experiments had the highest moving body speeds, especially experiment 4.

\begin{table}[h]
\centering
\caption{Error in Type 1 and Type 2 experimental texts for the MPU9255 IMU.}
\begin{tabular}{l l l}
\hline
\textbf{Experiment} & \textbf{Type 1 error (mm)} & \textbf{Type 2 error (mm)}\\
\hline
	1 & $16.7$ & $36.7$  \\ 
			
			2 & $28.4$ & $26.5$  \\ 
			
			3 & $25.3$ & $12.7$  \\ 
			
			4 & $12.0$ & $43.1$  \\ 
			
			5 & $16.1$ & $31.6$  \\
\hline
\end{tabular}
\label{table:results}
\end{table}

\begin{figure}
	\centering
	\includegraphics[width=0.8\linewidth]{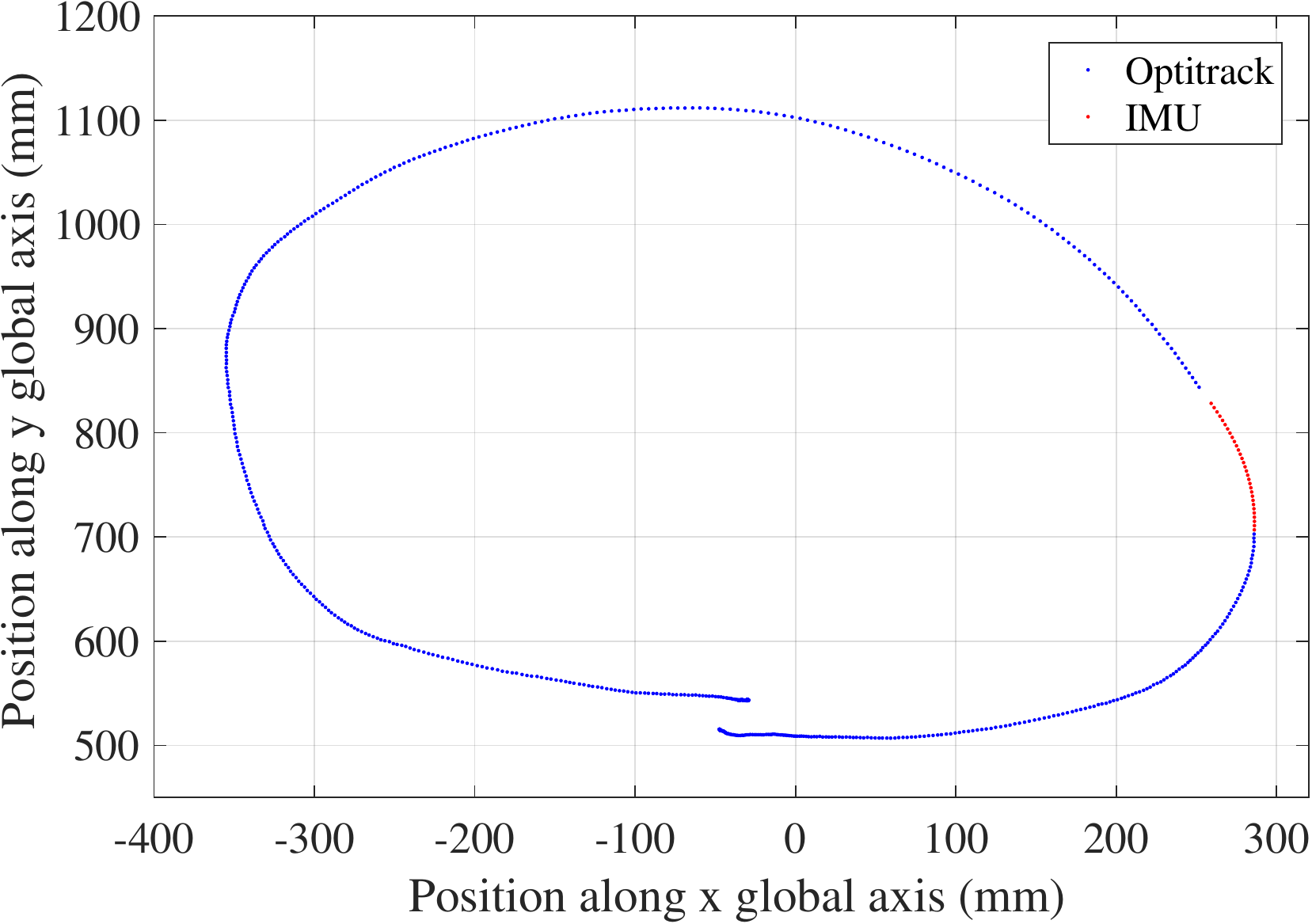}
	\caption{Optitrack position readings (blue) and IMU data position estimation (red)}
	\label{fig:circle}
\end{figure}

For the second test, roughly circular trajectory, the observed position estimation results are in Fig. \ref{fig:circle}. Again, the gap was placed on instants where the motion changes abruptly. We can observe that the motion is adequately tracked throughout both axes with an approximate error of 17.5 millimetres. This value is quite similar to the error obtained in previous tests.

The tests demonstrated that the gaps on the main motion tracking device (Optitrack) were no longer than 0.3 seconds. This time interval is sufficiently small to be properly compensated by the IMU data. Larger intervals of time accumulate significant error of position estimation from acceleration readings. The introduction of the IMU proves to be able to handle the motion tracking during these intervals. Considering a low-cost IMU and no prevalent noise filtering, these results are satisfactory as proof of concept.

\begin{figure}
	\centering
	\includegraphics[width=0.8\linewidth]{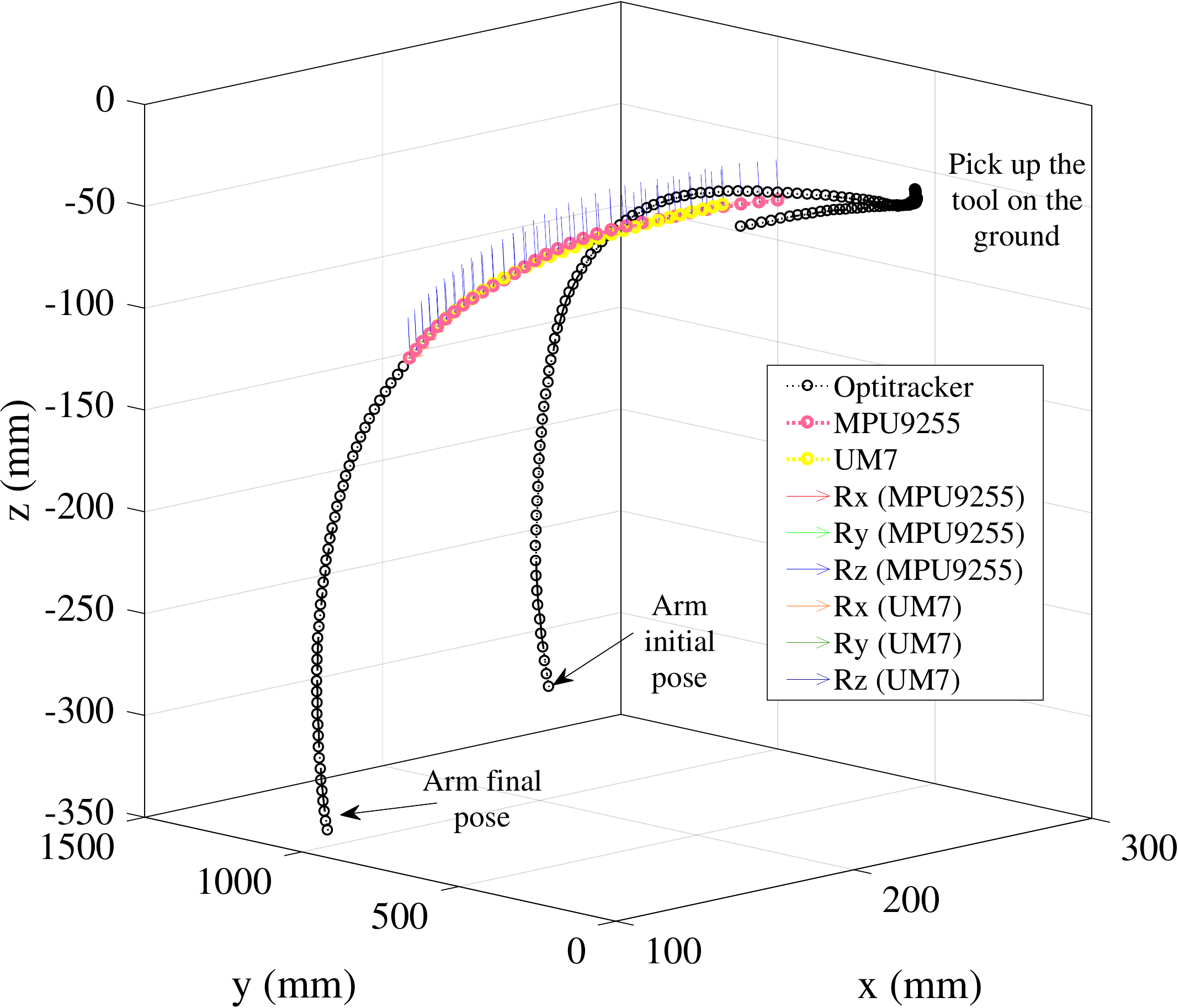}
	\caption{Use case position and orientation (only for IMUs) estimation. In the presence of occlusion, the IMUs present different performance, but both with the error in an acceptable range.}
	\label{fig:3d}
\end{figure}

\begin{figure}
	\centering
	\includegraphics[width=0.8\linewidth]{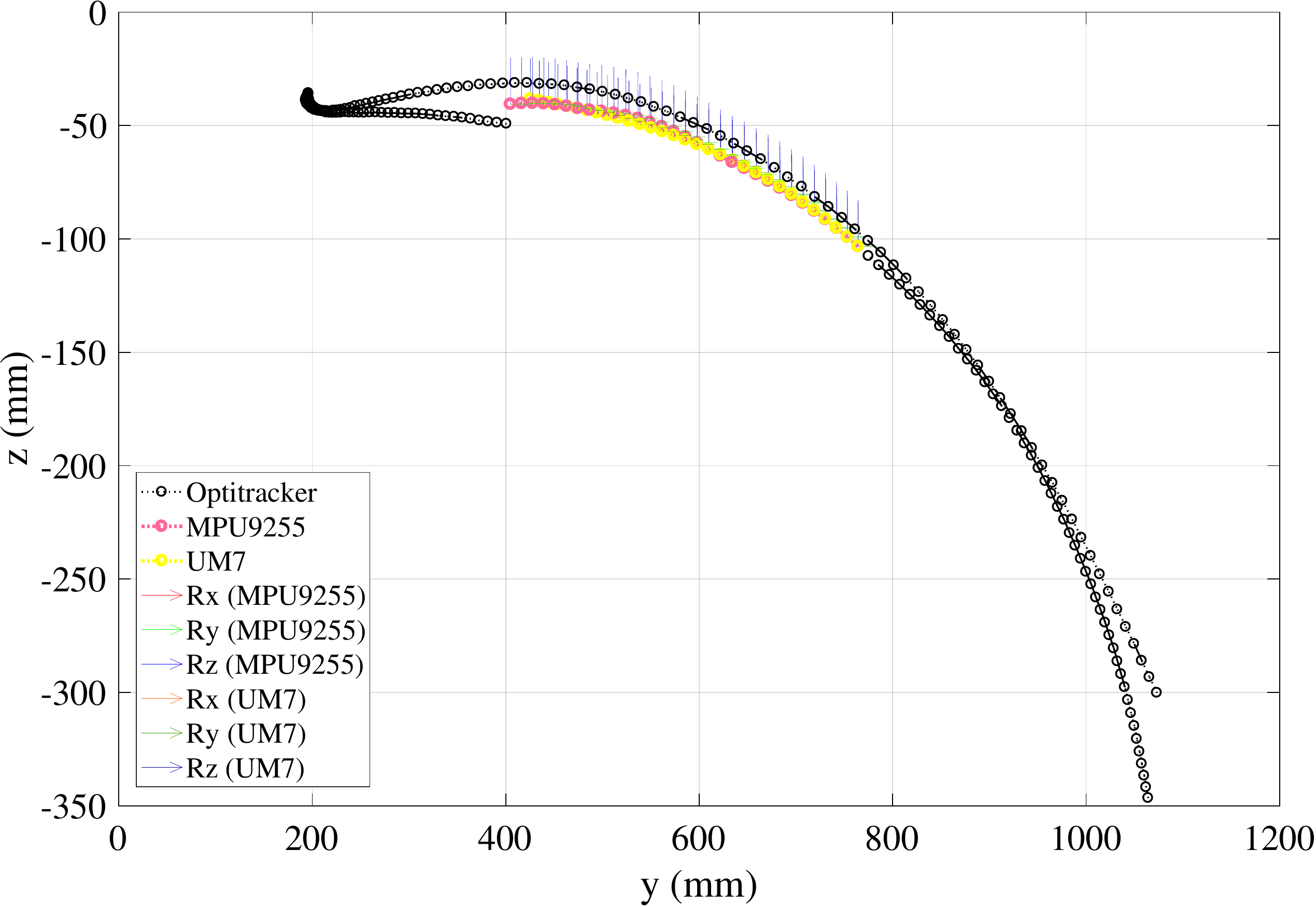}
	\caption{Use case position and orientation (only for IMUs) estimation in YZ frame. }
	\label{fig:2d}
\end{figure}

For the proposed use case, both IMUs were used to compare their performances. The occlusion took place during the operator’s forearm downward motion, which mainly corresponds to a motion in the global YZ frame. The error results, from both the specific axis and the global motion, are detailed on Table~~\ref{table:usecase_result}. The error obtained is in line with the error in the previous experiments. Fig. \ref{fig:3d} shows the arm positioning in space (including orientations) while Fig. \ref{fig:2d} shows the same data in YZ frame.

\begin{table}[h]
\centering
\caption{IMUs error comparison for the proposed use case.}
\begin{tabular}{l l l l}
\hline
\textbf{IMU} & \textbf{Z error (mm)} & \textbf{Y error (mm)} & \textbf{Total error (mm)}\\
\hline
	MPU9255 & $11.7$ & $14.8$ & $18.9$ \\ 
		
		UM7 & $11.4$ & $5.8$ & $12.8$ \\
\hline
\end{tabular}
\label{table:usecase_result}
\end{table}

Globally, analysing the results we can conclude that the benefit of using a more expensive IMU (UM7) is not significative in our application. The obtained error in the range of 10 to 30 millimetres is not affecting the use case functionality since we are dealing with bigger tolerances to avoid the collision between robot and human. In case the system is used in other applications, for example gesture recognition, the classification algorithms should be able to deal with such magnitude or error.

Comparing our results with the ones from similar studies, it can be concluded that in our proposed system the human tracking accuracy is in line with existing studies reporting the entire human body tracking. Our system presents the advantage of being reliable to occlusions. Even if it is for short periods of time, it is highly relevant in the human-robot interactive/collaborative process. On the other hand, it requires to use wearable sensors, which are less natural to interface than vision-based solutions.  

\section{Conclusion and future work}

The proposed algorithm for position estimation demonstrated relative positive results. Even under non-ideal testing conditions (occlusions), a sufficiently accurate position estimation with an error of approximately 15-17 millimetres was obtained using a low-cost IMU while tracking movements of over 700-800 millimetres. By adding rotation estimation using the orientation data provided by the gyroscope to update the rotation matrix from the body to the global frame, real-time pose estimation in the event of an occlusion was achieved. We concluded that the proposed online offset calculation had a critical positive impact on the usage of the selected low-cost IMU.

In future work, the main goal is to extend the application of our method to the whole human body to robustly estimate the full motion of a human. The integration of the sensors into smart textile/clothes is another research topic to study. We also aim to compare the performances of other IMUs from different price ranges. We expect to obtain a more accurate pose estimation and improve the quality of the readings, which in turn might extend the time interval where the IMU is able to estimate the pose of a human operator alone. It will be conducted a survey to evaluate the user acceptance of the proposed system while interfacing and share the workspace with a collaborative robot.

\section*{Acknowledgements}

This research was partially supported by Portugal 2020
project DM4Manufacturing POCI-01-0145-FEDER-016418
by UE/FEDER through the program COMPETE 2020, and
the Portuguese Foundation for Science and Technology COBOTIS project (PTDC/EME-
EME/32595/2017). This research is also sponsored by FEDER funds through the program COMPETE Programa Operacional Factores de Competitividade, and by national funds through FCT Funda\c{c}\~{a}o para a Ci\^encia e a Tecnologia under the project UIDB/00285/2020.

%% New version of the num-names style
\bibliographystyle{elsarticle-num-names}
%%\bibliography{sample.bib}

\end{document}